\documentclass[a4paper,fleqn]{cas-sc}
\def\degree{${}^{\circ}$}

\usepackage[numbers,sort&compress]{natbib}
\usepackage{array}

\makeatletter
\newcommand{\thickhline}{%
	\noalign {\ifnum 0=`}\fi \hrule height 2pt
	\futurelet \reserved@a \@xhline
}
\newcolumntype{"}{@{\hskip\tabcolsep\vrule width 1pt\hskip\tabcolsep}}
\makeatother

\def\tsc#1{\csdef{#1}{\textsc{\lowercase{#1}}\xspace}}
\tsc{WGM}
\tsc{QE}

\begin{document}
\let\WriteBookmarks\relax
\def\floatpagepagefraction{1}
\def\textpagefraction{.001}

\shorttitle{Gait-Transformer}    

\shortauthors{C.Z.}  

\title [mode = title]{Spatial Transformer Network on Skeleton-based Gait Recognition}

%

\author[1]{Cun Zhang}[type=author,style=chinese]

\cormark[1]


\ead{zhangcundejia@163.com}


\affiliation[1]{organization={ South China University of Technology},
            city={Guang Zhou},
            state={Guang Dong},
            country={China}}
\author{Xing-Peng Chen}[type=author,style=Chinese]
\author[1]{Guo-Qiang Han}[type=author,style=Chinese]
\author{Xiang-Jie Liu}[type=author,style=Chinese]

\cortext[1]{Corresponding author}



\begin{abstract}
Skeleton-based gait recognition models usually suffer from the robustness problem, as the Rank-1 accuracy varies from 90\% in normal walking cases to 70\% in  walking with coats cases. In this work, we propose a state-of-the-art robust skeleton‐based gait recognition model called Gait-TR, which is based on the combination of spatial transformer frameworks and temporal convolutional networks. Gait-TR achieves substantial improvements over other skeleton‐based gait models with higher accuracy and better robustness on the well-known gait dataset CASIA‐B. Particularly in walking with coats cases, Gait-TR got a $\sim$ 90
\% Rank-1 gait recognition accuracy rate, which is higher than the best result of silhouette-based models, which usually have higher accuracy than the silhouette‐based gait recognition models. Moreover, our experiment on CASIA‐B shows that the spatial transformer can extract gait features from the human skeleton better than the widely used graph convolutional network.

\end{abstract}


\begin{highlights}
\item 
We propose a new skeleton‐based gait recognition model called Gait-TR, which for the first time applies the spatial transformer framework for skeleton‐based gait recognition. 
\item 
Gait-TR achieves state-of-the-art results on the CASIA-B dataset, compared to existing skeleton-based gait recognition models. Especially in walking with coat cases, Gait-TR is better than both existing skeleton-based and silhouette-based gait recognition models.
\item 
Our experiment on CASIA‐B shows that the spatial transformer can extract gait features from the human skeleton better than the graph convolutional network.
\item 
The proposed model can be faster with fewer parameters by reducing the model layers or gait sequence length, while the accuracy decreases a few (4-6\%). The faster inference speed, higher accuracy, and better robustness of our model make gait recognition a step closer to the applications of gait recognition in the wild.  
\end{highlights}


\maketitle



\section{Introduction}\label{intro}
Biometrics technology uses various physiological characteristics, such as faces, fingerprints, DNA, and iris, to identify or recognize a person. However, most of them require his or her cooperation, e.g. taking a facial picture in high resolution or fingerprints by a fingerprinting technician. Gait, a person's pattern of walking, is one of the biometric modalities that can be collected easily even using a low-resolution camera over a long distance. Also, a person's gait pattern is hard to fake. Therefore, gait has been one of the most important biometrics technologies  widely used in video surveillance systems.  

While gait can be captured by different devices, such as video cameras or motion sensors, we focus on video-based gait recognition in this work. The inputs of most video-based gait recognition algorithms are human silhouette sequences  (silhouette-based gait recognition) or human skeleton sequences  (skeleton-based gait recognition) which are detected from people walking videos. The performance of gait recognition models can be sensitive to two factors: original gait diversity from the scenes where gait videos are captured, and the human body silhouette segmentation (or skeleton detection) methods. For the first one, people may be walking with coats or carrying items, the video cameras could be in different views, there could also be clutter in the scenes, etc. The second factor comes from the data preprocessing stage of gait recognition models, whose effects can be reduced by the recent developments in human body silhouette segmentation (and human body skeleton detection) research. All these complex factors make gait recognition more challenging. 

In the past two decades, lots of research studies have been conducted to solve challenges in gait recognition\cite{wan2018survey,kusakunniran2020review,rida2019robust,sepas2021deep}. Several gait datasets were collected, including the well-known CASIA-B\cite{CASIA} and OU-MVLP\cite{MVLP}. Some challenging factors for gait recognition, such as carrying, dressing, and different views, are considered in these gait datasets.  Also, plants of gait recognition models were developed, ranging from non-deep methods to the recent deep learning-based networks. Recently, the most popular two classes of gait recognition models are the appearance-based (silhouette-based) models and model-based models, which use human silhouettes and human pose as input respectively.  

The silhouette-based models were studied a lot and achieved state-of-the-art results in most gait datasets by the introduction of several significant methods. In 2016, K.Shiraga et al. proposed a gait recognition model named GEINet using a convolutional neural network, which yields two times better accuracy better than past models. GEINet \cite{shiraga2016geinet} was one of the first groups of silhouette-based models using deep learning-based networks. Since then, the performance of silhouette-based models has increased sharply. Most new models focused on extracting both the spatial information and temporal information of a gait sequence. GaitSet\cite{chao2021gaitset,chao2019gaitset} is the first silhouette-based model which regards gait as a set to extract temporal information. Then B.Lin et al. used multiple-temporal-scale 3D CNN to combine both small and large temporal scales spatial-temporal features\cite{lin2020gait}. Recently, T. Chai et al. developed the state-of-the-art silhouette-based model Vi-GaitGL \cite{chai2021silhouette} which uses the multi-task learning method with GaitGL as the backbone.

Compared with silhouette-based models, skeleton-based models have several advantages. Firstly, human skeletons can be extracted from images or videos more easily. Secondly, human skeletons consist of several key points, that are convenient for data storage and transformation. Thirdly, human skeletons are free from redundant features such as hairstyle, which makes the skeleton-based model more robust. Great improvement in skeleton-based models has been observed in recent years. In 2019, R.Liao et al. proposed the PoseGait\cite{liao2020model} which uses estimated human 3D poses as inputs, while a simple CNN was applied to get  Spatio-temporal features.  In 2021, T.Teepe  et al. proposed the GaitGraph\cite{teepe2021gaitgraph} which uses ResGCN\cite{song2020ResGCN} as basic blocks. The ResGCN is composed of a graph convolutional network followed by a temporal convolutional network.  In the same year,  the state-of-the-art skeleton-based model Gait‐D\cite{gao2022gait} was proposed which applies a similar network as the gait feature extractor. 
    
However, the performance of most existing skeleton-based models is worse than that of silhouette-based models. To get better spatial-temporal features from skeleton gait sequence, in this work, we propose a new skeleton-based gait recognition model, which applies the spatial transformer network\cite{plizzari2021spatial} as the spatial feature extractor, and the temporal convolutional network as the temporal feature extractor.

The main contributions of this work can be summarized as follows:
\begin{itemize}
\item 
We propose a new skeleton‐based gait recognition model called Gait-TR, which for the first time applies the spatial transformer framework for skeleton‐based gait recognition. 
\item 
Gait-TR achieves state-of-the-art results on the CASIA-B dataset, compared to existing skeleton-based gait recognition models. Especially in walking with coat cases, Gait-TR is better than both existing skeleton-based and silhouette-based gait recognition models.
\item 
Our experiment on CASIA‐B shows that the spatial transformer can extract gait features from the human skeleton better than the graph convolutional network.
\item 
The proposed model can be faster with fewer parameters by reducing the model layers or gait sequence length, while the accuracy decreases a few (4-6\%). The faster inference speed, higher accuracy, and better robustness of our model make gait recognition a step closer to the applications of gait recognition in the wild. 
\end{itemize}

\section{Related Work}\label{work}
In this section, we provide a brief overview of the two important groups of gait recognition methods: appearance-based methods and model-based methods. 
 As the human skeleton is the input of our proposed model, we briefly introduce human pose estimation at the end of this section.

\subsection{Gait Recognition}\label{review}
 \textbf{Appearance-based methods.} The appearance-based gait recognition methods identify different objects by features extracted from the appearance of individuals. The raw inputs of appearance-based methods are human silhouettes. Therefore, a data preprocessing step is required to segment human silhouettes from videos or image sequences. One of the popular gait features is gait energy image(GEI) which is the average of sequential silhouettes over one gait period. GEI-based methods (such as GEI+PCA) achieved good accuracy and were easy to be calculated, thus GEI-based methods were well studied in the early stage of appearance-based gait recognition research. However, the temporal average operator in  GEI leads to the missing of some temporal information. Also, large performance variations from view and orientation changes were observed.
 
 In recent years, appearance-based gait recognition research mainly focused on the application of deep neural network architectures and used the whole sequence of human silhouettes as input.  These deep appearance-based methods achieved much better performance than the old methods. Various neural network frameworks have been used, including convolutional neural networks (CNNs)\cite{shiraga2016geinet,wu2016comprehensive}, Recurrent Neural Networks (RNNs)\cite{jun2020feature,hasan2020multi}, and Generative Adversarial Networks (GANs)\cite{hu2018gan,wang2019gan}. Moreover, recently several deep learning strategies were applied to improve the performance of gait recognition models, including self-supervised learning and multi-task learning.  In ref.\cite{chao2019gaitset}, H.Chao et al. regarded a gait sequence as a set consisting of independent gait frames, which could drop unnecessary sequential constraints. Their proposed model, GaitSet, achieves 96.1\% rank-1 recognition accuracy on the CASIA-B gait dataset under normal walking conditions (The gait recognition accuracy is calculated with identical-view excluded in this work unless otherwise stated). Moreover, GaitSet even got 85.0\% accuracy using only 7 frames.
 On the other hand, MT3D applies a multiple-temporal-scale 3D Convolutional Neural Network to extract both small and large temporal scales gait information. MT3D achieves state-of-the-art results with accuracy of 96.7\% and 93.0\%, under normal walking and walking with a bag condition, respectively.  The state-of-the-art appearance-based gait recognition model is Vi-GaitGL proposed by T.Chai et al in Ref.\cite{chai2021silhouette} with an average accuracy of 92.2\%.  Vi-GaitGL adopts multi-task Learning to view-specific gait recognition model by fitting view angle along with gait recognition. And GaitGL, which consists of global and local convolutional neural network blocks, is used as the backbone. Under the walking with coats condition, Vi-GaitGL achieves an accuracy of 87.2\%.

 \textbf{Model-based methods.}
 Model-based gait recognition method is defined gait recognition approach which uses an underlying mathematical construct modeling the body structures or local body movements to discriminate different gait styles. Compared with appearance-based methods, model-based methods are free from several noisy variations from human silhouettes in conditions such as clothing and
 carrying, making model-based methods focus on the gait dynamics. Therefore, model-based methods were thought to be more robust. However, the accuracy of model-based methods in most of the existing research is lower than that of appearance-based methods, which made model-based methods less popular. Ref.\cite{nixon1996earlest} is one of the easiest works about model-based methods. In Ref.\cite{nixon1996earlest}, M. S. Nixon et al. got gait features by applying a simple Fourier transform to the motion of legs. Then k-nearest neighbors algorithm was used to classify ten gait subjects. After that, many feature extraction methods were proposed by analyzing patterns in gait databases, which was very tedious. 
 
Developments of the deep neural network and human pose estimation methods led to a new stage of skeleton-based gait recognition research. In Ref.\cite{liao2020model}, R.Liao et al. proposed the PoseGait which is based on
human 3D poses extracted by the pose estimation model OpenPose\cite{cao2017openpose}.  Specially designed Spatio-temporal features, such as joint angle, limb length, and joint motion are used as input of a deep feature extractor composed of CNN layers. PoseGait achieved good performance in identical-view cases, while the
accuracy in cross-view cases is still less than that of appearance-based methods.

 More recently, with the Graph Convolutional Network\cite{zhang2019graph,kipf2016semi} applied as a better skeleton feature extractor, model-based methods got breakthroughs with better accuracy and robustness, such as GaitGraph and Gait-D. The GaitGraph, proposed by T.Teepe, is composed of multiple ResGCN blocks. And a better 2D human pose estimator,  HRNet, is applied.  Gait-D is the state-of-the-art model-based gait recognition method proposed in Ref.\cite{gao2022gait}. The network structure of Gait-D is similar to GaitGraph. While in Gait-D, the canonical polyadic decomposition algorithm is used to decompose features extracted from ST‐GCN\cite{yan2018spatial} blocks. The accuracy of Gait-D is close to the best result of appearance-based methods in the CASIA-B dataset.
 
\subsection{Human Pose Estimation}\label{review}
Human pose estimation is one of the most popular fundamental tasks in computer vision. Human pose estimation aims to localize human body parts and human body keypoints from images or videos. Information about the human body (parts, key points, or skeleton) extracted by human pose estimation could be used in a lot of applications such as human-computer interaction, virtual reality, and augmented reality. Therefore, a lot of research about human pose estimation has been conducted in academia,  for comprehensive reviews about human pose estimation see Ref.\cite{khan2018review,liu20182,zheng2020deep,gamra2021review}. The human pose estimation methods are categorized into single-person and multi-person settings, or 3D based and 2D based.  OpenPose\cite{cao2017openpose} and HRNet\cite{sun2019deep} are the two most popular human pose estimation methods. In this work, we use the SimDR$^\ast$-HRNet proposed in Ref.\cite{li20212d} for 2D human pose estimation.

\section{Method}\label{method}

\begin{figure}
	\centering
	\includegraphics[width=0.99\textwidth]{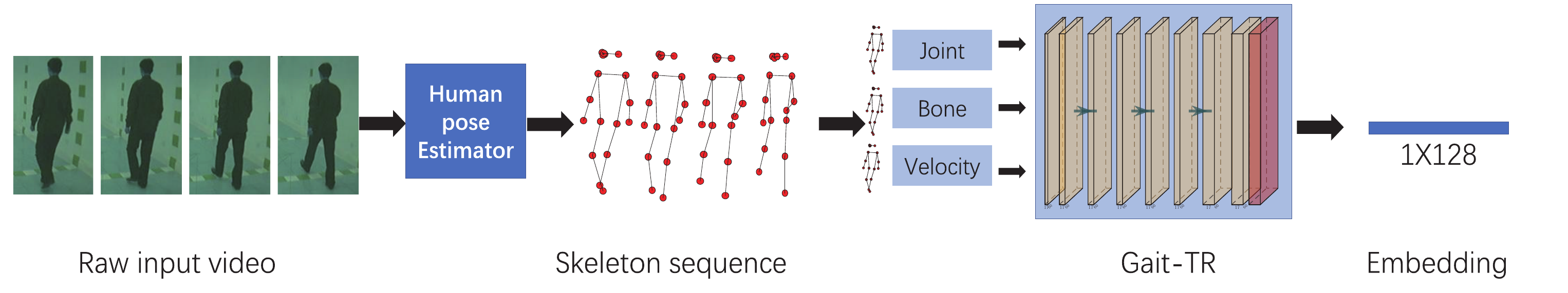}
	\caption{Pipeline of our framework}\label{fig1}
\end{figure}

In this part, we will illustrate our proposed framework for the skeleton-based gait recognition method. Fig.\ref{fig1} shows the pipeline of our framework. Firstly, we use a human pose estimator to extract skeleton sequences from the raw video. Secondly, we normalize the skeleton sequences and prepare different designed skeleton features(such as joints, bones, and velocities) as input channels. Finally, the Gait-TR processes with prepared input channels and outputs a 128 dimension embedding vector. In the inference phase, the Euclidean distances between the embedding vectors of two input videos are applied to distinguish different objects. 

Before going into detail, we introduce the most important part of our framework, namely, the spatial transformer.
\subsection{Spatial Transformer}\label{ST}
The transformer is the most popular neural network architecture in the past five years, proposed by A.Vaswani at el. in the paper "Attention is all you need"\cite{vaswani2017attention}. At first, the transformer was designed to replace RNN models widely used in natural language processing(NLP) and achieved state-of-the-art results in most of the NLP tasks\cite{wolf2020transformers,kalyan2021ammus,wolf2019huggingface,qiu2020pre}. Then the success of transformer architecture makes the transformer famous and be applied  in nearly all AI tasks, including computer vision\cite{han2020survey,ruan2022survey,liu2021swin}, biometrics\cite{sandouka2021transformers,zhong2021face}, music generation\cite{huang2018improved,huang2018music}, etc. 

The kernel of transformer architecture is the multi-head self-attention mechanism, which is described as follows. Given an input embedding $\textbf{x} \in \mathbb{R}^n$,  firstly, compute a query vector $\textbf{q}_h \in \mathbb{R}^{d_q}$, a key vector $\textbf{k}_h \in \mathbb{R}^{d_k}$, and a value vector $\textbf{v}_h \in \mathbb{R}^{d_v}$ by multiplying $\textbf{x}$ with the parameter matrix, $\textbf{W}^q_h \in \mathbb{R}^{n \times d_q}$, $\textbf{W}^k_h \in \mathbb{R}^{n \times d_k}$ and $\textbf{W}^v_h \in \mathbb{R}^{n \times d_v}$, respectively, for each head $i$ of the total $H$ heads. 
Then a scaled dot-product attention function is applied  to each query, key, and value:
\begin{eqnarray*}
	{\rm head}_h={\rm Attention}\left(\textbf{q}_h,\textbf{k}_h,\textbf{v}_h \right)={\rm softmax} \left(  \frac{\textbf{q}_h \textbf{k}_h^{\rm T}}{\sqrt{d_k}} \right) \textbf{v}_h
\end{eqnarray*}
Finally,  embedding vectors from  $h$ heads are concatenated and linear projected to final embedding $\textbf{z}\in  \mathbb{R}^{d_{model}}$:
\begin{eqnarray*}
	\textbf{z}={\rm Concat}({\rm head}_1,{\rm head}_2, \cdots,{\rm head}_H)W^o
\end{eqnarray*}
where $W^o \in  \mathbb{R}^{h*d_v \times d_{model}}$ is the projection matrix.

In this work, our inputs are human skeleton sequences: $X_{v}^t \in  \mathbb{R}^{C\times T \times V} $ for T frames, V joints, and C channels. Therefore, the spatial self-attention module of the spatial transformer proposed in Ref.\cite{plizzari2021spatial} is applied here.  In the spatial self-attention module, the attention functions contain correlations between the different nodes, that is:
\begin{eqnarray*}
	{\rm head}_{h}^t={\rm Attention}\left(\textbf{q}_h^t,\textbf{k}_h^t,\textbf{v}_h^t \right)=\sum_{j}{\rm softmax}_j \left(  \frac{\textbf{q}_{h,i} \textbf{k}_{h,j}^{\rm T}}{\sqrt{d_k}} \right) \textbf{v}_{h,j}
\end{eqnarray*}
All parameters in spatial self-attention are shared among different frames. In this work, we employ h=8 heads. For the dimension of query, key, and value vector, $d_q=d_k=f_k\times d_{model}$,$d_v=f_v \times d_{model}$, where $d_{model}$ is the output channel number of spatial self-attention block, $f_k$ and$f_v$ are fixed factors. 

\subsection{Data Preprocessing}\label{preprocess}
We use SimDR$^\ast$-HRNet as the 2D human pose estimator. The outputs of SimDR$^\ast$-HRNet are coordinates of 17 human body joints which are the nose, left ear, right ear, etc. In the training phase, we randomly select continuous skeleton sequences from the total skeleton sequence of a gait video, while in the testing phase, total skeleton sequences are used. 

As multiple inputs (which are simple features eg. bones, velocities, etc.) have been shown to be useful in some human skeleton-based tasks\cite{song2020ResGCN,shi2019two}, here we imply multiple inputs to get better performance. Given raw human skeleton joints $X$, joint features include joint coordinates  $X[:,:,i]$ and joint coordinates related to the nose $X[:,:,i]-X[:,:,i_{nose}]$. For velocity features, we use the first and second-order frame differences  as  $X[:,t+1,:]-X[:,t,:]$, $X[:,t+2,:]-X[:,t,:]$. The bone feature is defined as  $X[:,:,i]-X[:,:,i_{adj}]$, where $i_{adj}$ denotes the adjacent joint of the i-th joint.
Finally, we concatenate these features as input of Gait-TR.

\subsection{Gait-TR}\label{network}
\begin{figure}
	\centering
	\includegraphics[width=0.80\textwidth]{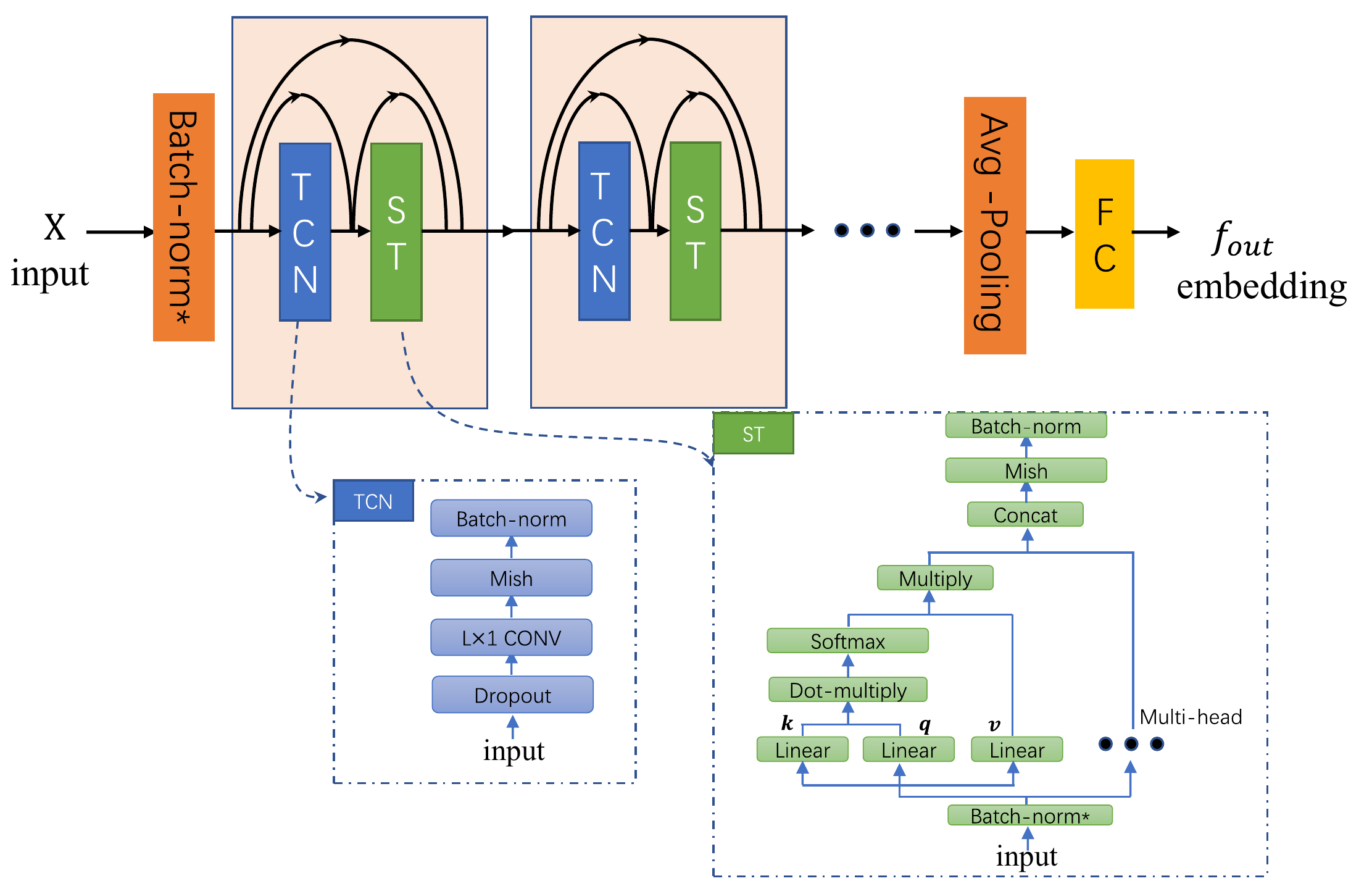}
	\caption{Structure of gait-TR.  TCN is the temporal convolutional network module, and ST is the spatial transformer module. FC denotes full connect layer. Batch-norm is BatchNorm2D for input $ X_{v}^t \in  \mathbb{R}^{C\times T \times V} $, while Batch-norm* denotes BatchNorm1D for input $ X_{v}^t \in  \mathbb{R}^{C*V\times T } $. }\label{fig2}
\end{figure}

Our proposed network Gait TRansformer (Gait-TR)  is constructed by stacking some basic blocks composed of a temporal convolutional network(TCN) module and a spatial transformer(ST)  module, shown in Fig.\ref{fig2}. Temporal convolutional network blocks are a plain convolutional network with kernel size $L$ along the temporal dimension, followed by the Mish activation function and batch normalization. Mish activation function is defined as $Mish(x)=x*tanh(softplus(x))$  proposed in Ref.\cite{misra2019mish}. Mish activation function and batch normalization are also used in the spatial transformer(ST) module. At the end of Gait-TR, an average pooling layer over temporal and spatial dimensions is used, and a full connect layer is applied to transform the dimension of features to the desired dimension.

The dense residual connection is used inside each TCN+ST block. The residual function is defined as:
\begin{equation*}
	H_{res}(x)=\left\{\begin{array}{ll}
		F(x)+x&{\rm size}(x)=={\rm size}(F(x)),\\
		F(x)+ {\rm Batchnorm}\left({\rm Mish}\left(Wx\right)\right)   & {\rm else}
	\end{array}\right.
\end{equation*}
where the last terms in the right equation are residual terms.

\section{Experimental Results}\label{experiments}
In this section, we evaluate the performance of the proposed Gait-TR on the gait dataset CASIA-B. First, we will introduce the details of the experiment, including the dataset, network structure, training setup, etc. Then we compare our result with both skeleton-based and silhouette-based gait
recognition methods.  Finally, we survey the Gait-TR with different setups.
\subsection{CASIA-B}
CASIA-B dataset is a famous large-scale multiple-view human gait dataset widely used in gait recognition research.  CASIA-B consists of 13,640 gait sequences from 124 persons. The view angle of CASIA-B ranges from 0 \degree to 180\degree  with 18\degree  increments. There are  10 gait sequences per view of each person, under three different walking conditions: 6 sequences in normal walking(NM), 2 sequences in carrying bag(BG), and 2 sequences in walking with coats(CL). Following the settings in most research, we use the first 24, 62, and 74 objects as train-set, denoted as small-sample(ST), medium-sample (MT), and large-sample (LT) respectively. In the inference phase, the first four sequences in NM condition are used as gallery set, the last two sequences in NM condition(NM \#5-6), two sequences in BG condition(BG \#1-2), and two sequences in CL condition (CL \#1-2) make three probe subsets.

\subsection{Implementation Details}
As said in previous sections, Gait-TR is composed of TCN+ST blocks.
Configuration of Gait-TR is shown in Tab.\ref{table1}, with output dimensions and numbers of parameters. Block0-Block3 are four stacked TCN+ST blocks with different channels. 

 \begin{table}[htbp]
 	\caption{Overview configuration of Gait-TR. The shape of input data is chosen to be $\left(10\times 60 \times 17\right)$.}\label{table1}
 	\begin{tabular}{c|c|c|c}
 		\thickhline
 		Block&Module & Output dimension  & Parameters\\
 		\hline
 		Multi-input	& input & $10\times 60 \times 17$ &-   \\
 		\hline
 		Data Norm & Batch Norm  & $10\times 60 \times 17$  &  -  \\
 		\hline
 		Block0	&  \multirow{4}{*}{TCN+ST} & $64\times 60 \times 17$   &  8,278 \\
 		Block1	&   & $64\times 60 \times 17$   &  49,760  \\
 		Block2	&   & $128\times 60 \times 17$   &  856,32 \\
 		Block3	&   & $256\times 60 \times 17$   &  335,104 \\
 		\hline
 		Avg-pooling	& pooling & $256\times 1 \times 1$   &  - \\
 		\hline
 		FC & Full connect  & $128\times 1 \times 1$   &  32,768 \\
 		\hline
 	\end{tabular}
 \end{table}
\textbf{Loss.} For the loss function, we imply the online mining batch-hard triple loss. For a sample triplet $(a,p,n)$ where, $a$ denotes an anchor, $p$ as a positive object of the same class as the anchor,  $n$ as a negative object, the triple loss is defined as:
\begin{eqnarray*}
	\mathcal{L}_{\rm triple}={\rm max}(d(f_a, f_p) - d(f_a, f_n) + {\rm margin}, 0)
\end{eqnarray*}
where $f_a$ denotes the feature vector of anchor, and $d(f_a, f_p)$ is the Euclidean distance between feature vectors of $a$ and $p$. In this work, the margin in triple loss is set to 0.3. Batch-hard means that for each $a$, we select the positive with the biggest distance $d(a, p)$ and the negative with the smallest distance $d(a, n)$ among the batch.

\textbf{Augment.} We apply several human gait data augment methods in the training phase. Firstly, we apply an inverse operator to the human skeleton by swapping the coordinates of the left parts and the right parts of a skeleton, eg. ${\rm Swap}(x_{\rm Lnose},x_{\rm Rnose})$. Gaussian noises are added to each joint, and the same gaussian noise is added to all joints in a gait sequence. Finally, we randomly select a continuous joint sequence with a length of 60.

\textbf{Training.} Adam optimizer is used with a weight decay of 2e-5.  Training data batches are sampled with batch size $(4,64)$, which means 4 persons and 64 gait sequences each.  We applied the three-phase 1-cycle learning rate schedule strategy, where initial, maximum, and final learning rates are set to 1e-5, 1e-3, and 1e-8, respectively. Finally, we train our model for 10k-30K iterations.

\subsection{Results and analysis} 
\begin{table}[htbp]
	\caption{Averaged rank-1 accuracies on CASIA-B dataset for skeleton-based methods, excluding identical-view cases. Results of  PoseGait, GaitGraph, Gait-D are also shown for comparison.}\label{table2}
	\begin{tabular}{c|c|c|c|c|c|c|c|c|c|c|c|c|c|c}
		\thickhline
		\multicolumn{3}{c|}{Gallery NM\#1-4}&\multicolumn{11}{|c|}{0\degree-180\degree} & mean\\
		\hline
		\multicolumn{3}{c|}{Probe}& 0\degree&18\degree&36\degree &54\degree &72\degree &90\degree &108 \degree &126\degree &144\degree &162\degree &180\degree & mean\\
		\hline
		\multirow{3}{*}{ST}& NM\#5-6&Gait-TR &72.2&77.4&77.5&79.6&76.7&76.7&76.8&78.2&76.0&71.8&64.0&75.2\\ \cline{2-15}
		& BG\#1-2&Gait-TR &60.7&65.9&65.5&70.0&61.5&64.3&65.2&66.5&66.3&63.7&53.7&63.9\\ \cline{2-15}
		& CL\#1-2&Gait-TR &56.9&61.2&61.8&63.7&62.7&61.5&62.6&63.8&59.2&59.8&48.3&60.1\\ \cline{2-15}
		\thickhline
		\multirow{9}{*}{MT}&\multirow{3}{*} {NM\#5-6}&PoseGait &55.3& 69.6 &73.9& 75.0& 68.0& 68.2& 71.1& 72.9& 76.1& 70.4& 55.4& 68.7\\
		&&Gait-D &87.7&92.5&93.6&\textbf{95.7}&93.3&92.4&92.8&93.4&90.6&88.6&87.3&91.6\\
		&&Gait-TR &\textbf{93.2}&\textbf{94.6}&\textbf{93.7}&93.1&\textbf{95.6}&\textbf{93.2} &\textbf{93.1}&\textbf{94.7}&\textbf{95.1}&\textbf{94.0}&\textbf{87.7}&\textbf{93.5}\\
		\cline{2-15}
		& \multirow{3}{*} {BG\#1-2}&PoseGait &35.3&47.2&52.4&46.9&45.5&43.9&46.1&48.1&49.4&43.6&31.1&44.5\\
		&&Gait-D &78.2&80.1&79.3&80.2&78.4&77.6&80.4&78.6&79.1&80.2&\textbf{76.5}&79.0\\
		&&Gait-TR &\textbf{87.1}&\textbf{88.7}&\textbf{89.4}&\textbf{91.1} &\textbf{87.1}&\textbf{88.6}&\textbf{89.3}&\textbf{90.8}&\textbf{92.9}&\textbf{88.5}&74.0&\textbf{88.0}\\
		\cline{2-15}
		& \multirow{3}{*} {CL\#1-2}&PoseGait &24.3&29.7&41.3&38.8&38.2&38.5&41.6&44.9&42.2&33.4&22.5&36.0\\
		&&Gait-D &73.2&71.7&75.4&73.2&74.6&72.3&74.1&70.5&69.4&71.2&66.7&72.0\\
		&&Gait-TR &\textbf{78.7}&\textbf{81.7}&\textbf{84.0}&\textbf{87.0} &\textbf{86.5}&\textbf{85.7}&\textbf{88.3}&\textbf{85.0}&\textbf{85.7}&\textbf{84.0}&\textbf{78.3}&\textbf{84.0}\\
		\thickhline
		\multirow{6}{*}{LT}&\multirow{2}{*} {NM\#5-6}&GaitGraph &85.3&88.5&91.0&92.5&87.2&86.5&88.4&89.2&87.9&85.9&81.9&87.7\\
		&&Gait-TR &\textbf{95.7}&\textbf{96.4}&\textbf{97.9}&\textbf{97.0}&\textbf{96.9} &\textbf{95.5}&\textbf{95.1}&\textbf{96.1}&\textbf{96.6}&\textbf{96.0}&\textbf{92.4}&\textbf{96.0}\\
		\cline{2-15}
		& \multirow{2}{*} {BG\#1-2}&GaitGraph &75.8&76.7&75.9&76.1&71.4&73.9&78.0&74.7&75.4&75.4&69.2&74.8\\
		&&Gait-TR &\textbf{90.9}&\textbf{92.4}&\textbf{91.4}&\textbf{93.2} &\textbf{91.9}&\textbf{90.2}&\textbf{91.4}&\textbf{93.9}&\textbf{93.9}&\textbf{92.7}&\textbf{82.9}&\textbf{91.3}\\
		\cline{2-15}
		& \multirow{2}{*} {CL\#1-2}&GaitGraph &69.6&66.1&68.8&67.2&64.5&62.0&69.5&65.6&65.7&66.1&64.3&66.3\\
		&&Gait-TR &\textbf{86.7}&\textbf{88.2}&\textbf{88.4}&\textbf{89.7} &\textbf{91.1}&\textbf{90.7}&\textbf{93.2}&\textbf{93.8}&\textbf{93.2}&\textbf{91.2}&\textbf{83.6}&\textbf{90.0}\\
		\hline
	\end{tabular}
\end{table}

\begin{table}[htbp]
	\caption{Averaged rank-1 accuracies on CASIA-B dataset, compared with silhouette-based methods, including  GaitSet,  MT3D, Vi-GaitGL.}\label{table3}
	\begin{tabular}{c|c|c|c|c|c|c|c|c|c|c|c|c|c|c}
		\thickhline
		\multicolumn{3}{c|}{Gallery NM\#1-4}&\multicolumn{11}{|c|}{0\degree-180\degree} & mean\\
		\hline
		\multicolumn{3}{c|}{Probe}& 0\degree&18\degree&36\degree &54\degree &72\degree &90\degree &108 \degree &126\degree &144\degree &162\degree &180\degree & mean\\
		\hline
		\multirow{12}{*}{ST}&\multirow{4}{*} {NM\#5-6}&GaitSet &71.6&\textbf{87.7}&\textbf{92.6}&89.1&\textbf{82.4}&\textbf{80.3}&\textbf{84.4}&\textbf{89.0}&89.8&82.9&66.6&\textbf{83.3}\\
		&&MT3D &71.9&83.9&90.9&\textbf{90.1}&81.1&75.6&82.1&89.0&\textbf{91.1}&\textbf{86.3}&\textbf{69.2}&82.8\\
		&&Vi-GaitGL &70.7&83.6&89.0&89.1&78.5&71.8&79.6&86.1&88.8&84.7&66.5&80.7\\
		&&Gait-TR&\textbf{72.2}&77.4&77.5&79.6&76.7&76.7&76.8&78.2&76.0&71.8&64.0&75.2\\
		\cline{2-15}
		& \multirow{4}{*} {BG\#1-2}&GaitSet &64.1&76.4&81.4&82.4&\textbf{77.2}& \textbf{71.8}&\textbf{75.4}&80.8&81.2&75.7&59.4&\textbf{75.1}\\
		&&MT3D &\textbf{64.5}&\textbf{76.7}&\textbf{82.8}&\textbf{82.8}&73.2&66.9&74.0&\textbf{81.9} &\textbf{84.8}&\textbf{80.2}&\textbf{63.0}&74.0\\
		&&Vi-GaitGL &64.2&75.0&82.6&81.5&70.2&63.9&70.4&77.8&81.0&77.6&58.3&72.9\\
		&&Gait-TR &60.7&65.9&65.5&70.0&61.5&64.3&65.2&66.5&66.3&63.7&53.7&63.9\\
		\cline{2-15}
		& \multirow{4}{*} {CL\#1-2}&GaitSet &36.4&49.7&54.6&49.7&48.7&45.2&45.5&48.2&47.2&41.4&30.6&45.2\\
		&&MT3D &46.6&61.6&66.5&63.3&57.4&52.1&58.1&58.9&58.5&57.4&41.9&56.6\\
		&&Vi-GaitGL &50.8&64.3&\textbf{68.6}&\textbf{67.1}&60.4&54.2&59.6&\textbf{63.9} &\textbf{62.9}&\textbf{59.9}&41.5&59.4\\
		&&Gait-TR &\textbf{56.9}&\textbf{61.2}&61.8&63.7&\textbf{62.7}&\textbf{61.5} &\textbf{62.6}&63.8&59.2&59.8&\textbf{48.3}&\textbf{60.1}\\
		\thickhline
		\multirow{12}{*}{MT}&\multirow{4}{*} {NM\#5-6}&GaitSet &89.7&\textbf{97.9}&98.3&\textbf{97.4}&92.5&90.4&93.4&97.0&\textbf{98.9}&\textbf{95.9}&86.6&94.3\\
		&&MT3D &91.9&96.4&\textbf{98.5}&95.7&93.8&90.8&93.9&97.3&97.9&95.0&86.8&\textbf{94.4}\\
		&&Vi-GaitGL &90.8&95.9&97.7&95.9&93.3&91.5&\textbf{94.4}&\textbf{97.3}&97.3&95.4&86.9&94.2\\
		&&Gait-TR &\textbf{93.2}&94.6&93.7&93.1&\textbf{95.6}&\textbf{93.2}&93.1&94.7&95.1&94.0&\textbf{87.7}&93.5\\
		\cline{2-15}
		& \multirow{4}{*} {BG\#1-2}&GaitSet &79.9&89.8&91.2&86.7&81.6&76.7&81.0&88.2&90.3&88.5&73.0&84.3\\
		&&MT3D &86.7&92.9&\textbf{94.9}&92.8&88.5&82.5&87.5&92.5&95.3&92.9&81.2&89.8\\
		&&Vi-GaitGL &83.6&\textbf{92.9}&94.7&\textbf{93.1}&\textbf{89.4}&83.6&88.6& \textbf{93.6}&\textbf{96.1}&\textbf{93.3}&\textbf{81.5}&\textbf{90.0}\\
		&&Gait-TR &\textbf{87.1}&88.7&89.4&91.1&87.1&\textbf{88.6}&\textbf{89.3}&90.8&92.9&88.5&74.0&88.0\\
		\cline{2-15}
		& \multirow{4}{*} {CL\#1-2}&GaitSet &52.0&66.0&72.8&69.3&63.1&61.2&63.5&66.5&67.5&60.0&45.9&62.5\\
		&&MT3D &67.5&81.0&85.0&80.6&75.9&69.8&76.8&81.0&80.8&73.8&59.0&75.6\\
		&&Vi-GaitGL &71.2&\textbf{86.5}&\textbf{90.9}&89.0&83.9&77.2&84.8&\textbf{89.1}&\textbf{88.6} &81.0&63.7&82.3\\
		&&Gait-TR &\textbf{78.7}&81.7&84.0&87.0&\textbf{86.5}  &\textbf{85.7}&\textbf{88.3} &85.0&85.7&\textbf{84.0}&\textbf{78.3}&\textbf{84.0}\\
		\thickhline
		\multirow{12}{*}{LT}&\multirow{4}{*} {NM\#5-6}&GaitSet &91.1&\textbf{99.0}&\textbf{99.9}&\textbf{97.8}&95.1&94.5&\textbf{96.1} &98.3&99.2&98.1&88.0&96.1\\
		&&MT3D &95.7&98.2&99.0&97.5&95.1&93.9&96.1&\textbf{98.6}&\textbf{99.2}&\textbf{98.2}&92.0&\textbf{96.7}\\
		&&Vi-GaitGL &93.7&96.9&98.6&97.4&95.5&93.9&97.3&98.6&98.6&97.7&89.7&96.2\\
		&&Gait-TR &\textbf{95.7}&96.4&97.9&97.0&\textbf{96.9}&\textbf{95.5}&95.1&96.1&96.6&96.0&\textbf{92.4}&96.0\\
		\cline{2-15}
		& \multirow{4}{*} {BG\#1-2}&GaitSet &86.7&94.2&95.7&93.4&88.9&85.5&89.0&91.7&94.5&95.9&83.3&90.8\\
		&&MT3D &\textbf{91.0}&\textbf{95.4}&\textbf{97.5}&94.2&92.3&86.9&91.2&95.6&97.3&96.4&\textbf{86.6}&\textbf{93.0}\\
		&&Vi-GaitGL &89.6&94.5&95.6&\textbf{95.2}&\textbf{93.2}&87.3&\textbf{91.7}&\textbf{95.9} &\textbf{97.8}&\textbf{96.1}&85.5&92.9\\
		&&Gait-TR &90.9&92.4&91.4&93.2&91.9&\textbf{90.2}&91.4&93.9&93.9&92.7&82.9&91.3\\
		\cline{2-15}
		& \multirow{4}{*} {CL\#1-2}&GaitSet &59.5&75.0&78.3&74.6&71.4&71.3&70.8&74.1&74.6&69.4&54.1&70.3\\
		&&MT3D &76.0&87.6&89.8&85.0&81.2&75.7&81.0&84.5&85.4&82.2&68.1&81.5\\
		&&Vi-GaitGL &81.2&\textbf{92.4}&\textbf{94.9}&\textbf{93.3}&87.8&82.1&87.4&89.8&90.2&87.9&72.5&87.2\\
		&&Gait-TR &\textbf{86.7}&88.2&88.4&89.7&\textbf{91.1}&\textbf{90.7}&\textbf{93.2}&\textbf{93.8}&\textbf{93.2}&\textbf{91.2}&\textbf{83.6}&\textbf{90.0}\\
		\hline
	\end{tabular}
\end{table}
\textbf{Comparison with skeleton-based methods.} In Tab.\ref{table2}, we show the average rank-1 accuracies on CASIA-B dataset of our Gait-TR, under different conditions, alongside the existing skeleton-based gait recognition methods, including  PoseGait, Gait-D, and GaitGraph. Tab.\ref{table2} clearly shows that our model Gait-TR achieves state-of-the-art performance under most of the cross-view and probe conditions. Firstly in LT cases, the largest improvement happens under the CL situation, where the rank-1 accuracy of Gait-TR is 90\% which is  23.7\% larger than that of  GaitGraph. In the NM and BG situations, our average rank-1 accuracies are 96.0\% and 91.3\%, and the improvements over that of GaitGraph are  8.3\% to 16.5\%.  Then in MT cases, a large increase of average accuracies is achieved under BG and CL situations,  9\% and 12\%, compared to that of Gait-D. A small improvement of about 2\% is got under NM situation.
Finally, for the first time, we calculate the  rank-1  accuracies under the ST sample setting, while the mean rank-1 accuracies are 75.2\%, 63.9\%, and 60.1\% for NM, BG, and CL probe situations, respectively. 

The accuracies of Gait-TR vary less under different probe situations, compared to  Gait-D and GaitGraph, which means that our model has better robustness against probe condition changes such as bagging and clothing.
In addition, it can also be observed from Tab.\ref{table2} that accuracy drops a lot in all conditions, from 4\% to 14\%.
A similar drop in accuracy happens in other models, however, with a smaller gap. 

\textbf{Comparison with silhouette-based methods.} We compare the result of Gait-TR with that of the state-of-the-art silhouette-based gait models, including  GaitSet,  MT3D, Vi-GaitGL, shown in Tab.\ref{table3}. Firstly, under LT cases, our rank-1 accuracies of Gait-TR is bigger than the best by 3\%, in the CL situation. Meanwhile, the accuracies in NM  and BG are very close to those of the best silhouette-based methods, only 0.7\% and 1.7\% less than that of the best silhouette-based methods.  Performances in MT cases are similar to that in the LT cases. However, in ST cases, the accuracy of Gait-TR drops larger than the accuracy of these silhouette-based gaits, which means that  Gait-TR needs more gait data to get good enough performance. In ST cases, the performance with CL\#1-2 probe is still better than silhouette-based methods.
 
 \begin{table}[htbp]
 	\caption{ Mean Rank-1 accuracy, number of parameters and FLOPs of Gait-TR-s, along with other models including Gait-TR, GaitSet and GaitGraph.  The FLOPs are calculated using gait sequences of 60 frames.  }\label{table-small}
 	\begin{tabular}{c|c|c|c|c|c}
 		\thickhline
 		Model& {NM\#5-6} & {BG\#1-2}  & {CL\#1-2}&Parameter& FLOPs \\
 		\hline
 		GaitGraph &87.7&74.8&66.3&0.32M&0.28G\\
 		GaitSet &96.1&90.8&70.3&2.59M&13.02G\\
 		Gait-TR-s &92.2&86.2&85.3&0.16M&0.29G\\
 		Gait-TR &96.0&91.3&90.0&0.51M&0.98G\\
 		\hline
 	\end{tabular}
 \end{table}

\textbf{Smaller model.}
To get faster inference speed, we propose a model with fewer parameters, named  Gait-TR-s, whose structure is similar to Gait-TR, with the last TCN+ST block removed from Gait-TR. The performance (including rank-1 accuracy, number of parameters, and FLOPs) of Gait-TR-s is shown in Tab.\ref{table-small}, compared with other models.  The mean rank-1 accuracy of Gait-TR-s is lower than that of Gait-TR by 4\%-5\%.   Parameters and FlOPs of Gait-TR-s are 0.16M and 0.29GFlOPs, respectively, which are 2/3 less than that of Gait-TR. Silhouette-based methods (eg, GaitSet) need more parameters and FLOPs than skeleton-based methods. The faster inference speed and fewer parameters of skeleton-based methods provide other evidence to support the opinion that skeleton-based methods are more suitable for practical gait recognition.

\begin{figure}
	\centering
	\includegraphics[width=0.5\textwidth]{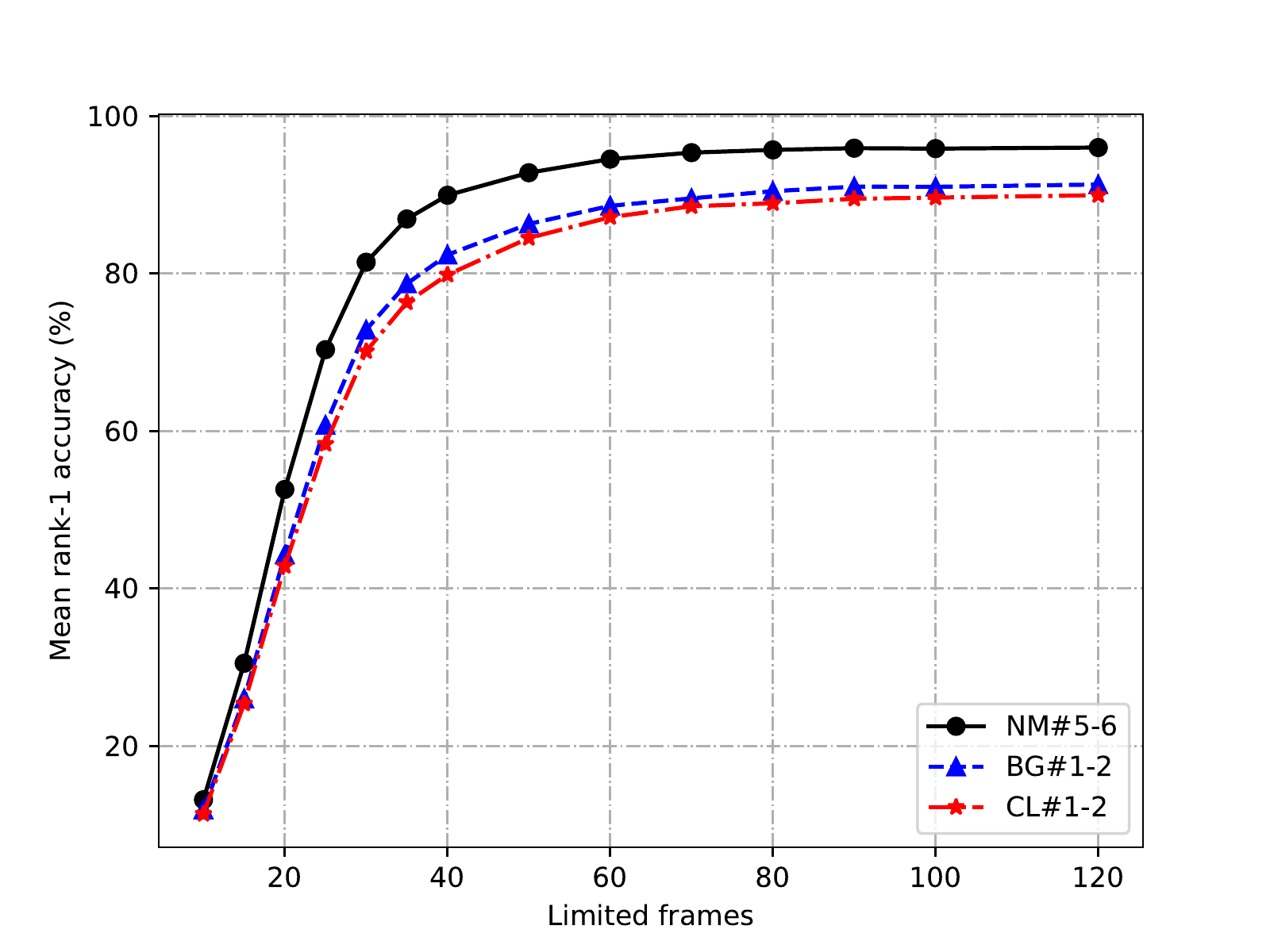}
	\caption{ Mean Rank-1 accuracy with limited inference frames. }\label{fig_frame}
\end{figure}

\textbf{Limited inference frame.}
In the practical application of gait recognition, the total number of frames in which a target is walking could be limited. Therefore, we test our model Gait-TR on limited frames of gait sequences. The gait sequences for inference are continuous gait sequences with length $T$. Fig.\ref{fig_frame} shows the mean ran-1 accuracy vs different sequences length for different probe conditions, under the LT sample set. The accuracies decrease sharply as frame length decreases from 50, which is twice a common gait cycle, 25. This indicates that our Gait-TR depends on the long frame feature of a gait sequence.  To get an accuracy large than 80\% under CL condition, the length of gait sequences need to be longer than 40. 

\textbf{Spatial Transformer vs Graph Convolutional Network.} 
Graph Convolutional Network(GCN) is a widely used spatial feature extractor for human skeleton sequences. Here we compare the spatial feature extractor of our Gait-TR, Spatial Transformer(ST), with GCN. We replace the ST module in Gait-TR with GCN, and name the resulting model as  Gait-GCN. Tab.\ref{table-gcn} shows the performance of Gait-TR and Gait-GCN. The accuracy of Gait-TR is higher than Gait-GCN by 2\% to 3\% with a similar inference speed. This result implies that ST can be a better spatial feature extractor than GCN in skeleton-based gait recognition.

 \begin{table}[htbp]
	\caption{Comparison between Gait-TR and Gait-GCN under LT sample condition.}\label{table-gcn}
	\begin{tabular}{c|c|c|c|c|c}
		\thickhline
		model& {NM\#5-6} & {BG\#1-2}  & {CL\#1-2}&parameters& FLOPs \\
		\hline
		Gait-TR &96.0&91.3&90.0&0.513M&0.976G\\
		Gait-GCN &94.5&88.8&87.1&0.482M&0.937G\\
		\hline
	\end{tabular}
\end{table}

\section{Conclusion}\label{conclusion}
In this work, we investigated, for the first time, the spatial transformer framework in skeleton-based gait recognition models. Our proposed model gait-TR achieves state-of-the-art results on the CASIA-B dataset compared to current skeleton-based models. Especially in walking with coats cases, the proposed model is even better than the existing silhouette-based models. Our experiment on CASIA‐B also shows that spatial transformer can extract gait features from the human skeleton better than the graph convolutional network.

In real-world scenarios, most silhouette extraction methods are more complex and slower than skeleton detection methods. Compared to silhouette-based models which need silhouette extraction in the data preprocessing step, skeleton-based models can do better in practical applications. However, in past works, the performance of skeleton-based models was worse than the performance of silhouette-based models. Therefore the better performance of skeleton-based than silhouette-based models in our work, although only in the walking with coats cases, shows the potential of skeleton-based models for higher accuracy and better robustness. Our proposed state-of-the-art skeleton-based gait recognition model makes gait recognition a step closer to the applications of gait recognition in the wild.   

As gait-TR is a skeleton-based model, better skeleton sequences from a better human pose estimator are beneficial.
Also, Gait-TR requires gait sequences of long-frame, about twice a gait cycle, to get good performance. A temporal feature extractor better than the simple temporal convolutional network could be valuable for better performance and practical applications with faster inference speed.

\printcredits

\bibliographystyle{unsrt}

\bibliography{gait}

\begin{thebibliography}{10}

\bibitem{wan2018survey}
Changsheng Wan, Li~Wang, and Vir~V Phoha.
\newblock A survey on gait recognition.
\newblock {\em ACM Computing Surveys (CSUR)}, 51(5):1--35, 2018.

\bibitem{kusakunniran2020review}
Worapan Kusakunniran.
\newblock Review of gait recognition approaches and their challenges on view
  changes.
\newblock {\em IET Biometrics}, 9(6):238--250, 2020.

\bibitem{rida2019robust}
Imad Rida, Noor Almaadeed, and Somaya Almaadeed.
\newblock Robust gait recognition: a comprehensive survey.
\newblock {\em IET Biometrics}, 8(1):14--28, 2019.

\bibitem{sepas2021deep}
Alireza Sepas-Moghaddam and Ali Etemad.
\newblock Deep gait recognition: A survey.
\newblock {\em arXiv preprint arXiv:2102.09546}, 2021.

\bibitem{CASIA}
Shiqi Yu, Daoliang Tan, and Tieniu Tan.
\newblock A framework for evaluating the effect of view angle, clothing and
  carrying condition on gait recognition.
\newblock In {\em 18th International Conference on Pattern Recognition
  (ICPR'06)}, volume~4, pages 441--444. IEEE, 2006.

\bibitem{MVLP}
Noriko Takemura, Yasushi Makihara, Daigo Muramatsu, Tomio Echigo, and Yasushi
  Yagi.
\newblock Multi-view large population gait dataset and its performance
  evaluation for cross-view gait recognition.
\newblock {\em IPSJ Transactions on Computer Vision and Applications},
  10(1):1--14, 2018.

\bibitem{shiraga2016geinet}
Kohei Shiraga, Yasushi Makihara, Daigo Muramatsu, Tomio Echigo, and Yasushi
  Yagi.
\newblock Geinet: View-invariant gait recognition using a convolutional neural
  network.
\newblock In {\em 2016 international conference on biometrics (ICB)}, pages
  1--8. IEEE, 2016.

\bibitem{chao2021gaitset}
Hanqing Chao, Kun Wang, Yiwei He, Junping Zhang, and Jianfeng Feng.
\newblock Gaitset: Cross-view gait recognition through utilizing gait as a deep
  set.
\newblock {\em IEEE Transactions on Pattern Analysis and Machine Intelligence},
  2021.

\bibitem{chao2019gaitset}
Hanqing Chao, Yiwei He, Junping Zhang, and Jianfeng Feng.
\newblock Gaitset: Regarding gait as a set for cross-view gait recognition.
\newblock In {\em Proceedings of the AAAI conference on artificial
  intelligence}, volume~33, pages 8126--8133, 2019.

\bibitem{lin2020gait}
Beibei Lin, Shunli Zhang, and Feng Bao.
\newblock Gait recognition with multiple-temporal-scale 3d convolutional neural
  network.
\newblock In {\em Proceedings of the 28th ACM International conference on
  Multimedia}, pages 3054--3062, 2020.

\bibitem{chai2021silhouette}
Tianrui Chai, Xinyu Mei, Annan Li, and Yunhong Wang.
\newblock Silhouette-based view-embeddings for gait recognition under multiple
  views.
\newblock In {\em 2021 IEEE International Conference on Image Processing
  (ICIP)}, pages 2319--2323. IEEE, 2021.

\bibitem{liao2020model}
Rijun Liao, Shiqi Yu, Weizhi An, and Yongzhen Huang.
\newblock A model-based gait recognition method with body pose and human prior
  knowledge.
\newblock {\em Pattern Recognition}, 98:107069, 2020.

\bibitem{teepe2021gaitgraph}
Torben Teepe, Ali Khan, Johannes Gilg, Fabian Herzog, Stefan H{\"o}rmann, and
  Gerhard Rigoll.
\newblock Gaitgraph: graph convolutional network for skeleton-based gait
  recognition.
\newblock In {\em 2021 IEEE International Conference on Image Processing
  (ICIP)}, pages 2314--2318. IEEE, 2021.

\bibitem{song2020ResGCN}
Yi-Fan Song, Zhang Zhang, Caifeng Shan, and Liang Wang.
\newblock Stronger, faster and more explainable: A graph convolutional baseline
  for skeleton-based action recognition.
\newblock In {\em proceedings of the 28th ACM international conference on
  multimedia}, pages 1625--1633, 2020.

\bibitem{gao2022gait}
Shuo Gao, Jing Yun, Yumeng Zhao, and Limin Liu.
\newblock Gait-d: Skeleton-based gait feature decomposition for gait
  recognition.
\newblock {\em IET Computer Vision}, 16(2):111--125, 2022.

\bibitem{plizzari2021spatial}
Chiara Plizzari, Marco Cannici, and Matteo Matteucci.
\newblock Spatial temporal transformer network for skeleton-based action
  recognition.
\newblock In {\em International Conference on Pattern Recognition}, pages
  694--701. Springer, 2021.

\bibitem{wu2016comprehensive}
Zifeng Wu, Yongzhen Huang, Liang Wang, Xiaogang Wang, and Tieniu Tan.
\newblock A comprehensive study on cross-view gait based human identification
  with deep cnns.
\newblock {\em IEEE transactions on pattern analysis and machine intelligence},
  39(2):209--226, 2016.

\bibitem{jun2020feature}
Kooksung Jun, Deok-Won Lee, Kyoobin Lee, Sanghyub Lee, and Mun~Sang Kim.
\newblock Feature extraction using an rnn autoencoder for skeleton-based
  abnormal gait recognition.
\newblock {\em IEEE Access}, 8:19196--19207, 2020.

\bibitem{hasan2020multi}
Md~Mahedi Hasan and Hossen~Asiful Mustafa.
\newblock Multi-level feature fusion for robust pose-based gait recognition
  using rnn.
\newblock {\em Int. J. Comput. Sci. Inf. Secur.(IJCSIS)}, 18(1), 2020.

\bibitem{hu2018gan}
BingZhang Hu, Yan Gao, Yu~Guan, Yang Long, Nicholas Lane, and Thomas Ploetz.
\newblock Robust cross-view gait identification with evidence: A discriminant
  gait gan (diggan) approach on 10000 people.
\newblock 2018.

\bibitem{wang2019gan}
Yanyun Wang, Chunfeng Song, Yan Huang, Zhenyu Wang, and Liang Wang.
\newblock Learning view invariant gait features with two-stream gan.
\newblock {\em Neurocomputing}, 339:245--254, 2019.

\bibitem{nixon1996earlest}
Mark~S Nixon, John~N Carter, D~Cunado, Ping~S Huang, and SV~Stevenage.
\newblock Automatic gait recognition.
\newblock In {\em Biometrics}, pages 231--249. Springer, 1996.

\bibitem{cao2017openpose}
Zhe Cao, Tomas Simon, Shih-En Wei, and Yaser Sheikh.
\newblock Realtime multi-person 2d pose estimation using part affinity fields.
\newblock In {\em Proceedings of the IEEE conference on computer vision and
  pattern recognition}, pages 7291--7299, 2017.

\bibitem{zhang2019graph}
Si~Zhang, Hanghang Tong, Jiejun Xu, and Ross Maciejewski.
\newblock Graph convolutional networks: a comprehensive review.
\newblock {\em Computational Social Networks}, 6(1):1--23, 2019.

\bibitem{kipf2016semi}
Thomas~N Kipf and Max Welling.
\newblock Semi-supervised classification with graph convolutional networks.
\newblock {\em arXiv preprint arXiv:1609.02907}, 2016.

\bibitem{yan2018spatial}
Sijie Yan, Yuanjun Xiong, and Dahua Lin.
\newblock Spatial temporal graph convolutional networks for skeleton-based
  action recognition.
\newblock In {\em Thirty-second AAAI conference on artificial intelligence},
  2018.

\bibitem{khan2018review}
Naimat~Ullah Khan and Wanggen Wan.
\newblock A review of human pose estimation from single image.
\newblock In {\em 2018 International Conference on Audio, Language and Image
  Processing (ICALIP)}, pages 230--236. IEEE, 2018.

\bibitem{liu20182}
Yi~Liu, Ying Xu, and Shao-bin Li.
\newblock 2-d human pose estimation from images based on deep learning: a
  review.
\newblock In {\em 2018 2nd IEEE Advanced Information Management, Communicates,
  Electronic and Automation Control Conference (IMCEC)}, pages 462--465. IEEE,
  2018.

\bibitem{zheng2020deep}
Ce~Zheng, Wenhan Wu, Taojiannan Yang, Sijie Zhu, Chen Chen, Ruixu Liu, Ju~Shen,
  Nasser Kehtarnavaz, and Mubarak Shah.
\newblock Deep learning-based human pose estimation: A survey.
\newblock {\em arXiv preprint arXiv:2012.13392}, 2020.

\bibitem{gamra2021review}
Miniar~Ben Gamra and Moulay~A Akhloufi.
\newblock A review of deep learning techniques for 2d and 3d human pose
  estimation.
\newblock {\em Image and Vision Computing}, 114:104282, 2021.

\bibitem{sun2019deep}
Ke~Sun, Bin Xiao, Dong Liu, and Jingdong Wang.
\newblock Deep high-resolution representation learning for human pose
  estimation.
\newblock In {\em Proceedings of the IEEE/CVF Conference on Computer Vision and
  Pattern Recognition}, pages 5693--5703, 2019.

\bibitem{li20212d}
Yanjie Li, Sen Yang, Shoukui Zhang, Zhicheng Wang, Wankou Yang, Shu-Tao Xia,
  and Erjin Zhou.
\newblock Is 2d heatmap representation even necessary for human pose
  estimation?
\newblock {\em arXiv preprint arXiv:2107.03332}, 2021.

\bibitem{vaswani2017attention}
Ashish Vaswani, Noam Shazeer, Niki Parmar, Jakob Uszkoreit, Llion Jones,
  Aidan~N Gomez, {\L}ukasz Kaiser, and Illia Polosukhin.
\newblock Attention is all you need.
\newblock {\em Advances in neural information processing systems}, 30, 2017.

\bibitem{wolf2020transformers}
Thomas Wolf, Lysandre Debut, Victor Sanh, Julien Chaumond, Clement Delangue,
  Anthony Moi, Pierric Cistac, Tim Rault, R{\'e}mi Louf, Morgan Funtowicz,
  et~al.
\newblock Transformers: State-of-the-art natural language processing.
\newblock In {\em Proceedings of the 2020 conference on empirical methods in
  natural language processing: system demonstrations}, pages 38--45, 2020.

\bibitem{kalyan2021ammus}
Katikapalli~Subramanyam Kalyan, Ajit Rajasekharan, and Sivanesan Sangeetha.
\newblock Ammus: A survey of transformer-based pretrained models in natural
  language processing.
\newblock {\em arXiv preprint arXiv:2108.05542}, 2021.

\bibitem{wolf2019huggingface}
Thomas Wolf, Lysandre Debut, Victor Sanh, Julien Chaumond, Clement Delangue,
  Anthony Moi, Pierric Cistac, Tim Rault, R{\'e}mi Louf, Morgan Funtowicz,
  et~al.
\newblock Huggingface's transformers: State-of-the-art natural language
  processing.
\newblock {\em arXiv preprint arXiv:1910.03771}, 2019.

\bibitem{qiu2020pre}
Xipeng Qiu, Tianxiang Sun, Yige Xu, Yunfan Shao, Ning Dai, and Xuanjing Huang.
\newblock Pre-trained models for natural language processing: A survey.
\newblock {\em Science China Technological Sciences}, 63(10):1872--1897, 2020.

\bibitem{han2020survey}
Kai Han, Yunhe Wang, Hanting Chen, Xinghao Chen, Jianyuan Guo, Zhenhua Liu,
  Yehui Tang, An~Xiao, Chunjing Xu, Yixing Xu, et~al.
\newblock A survey on visual transformer.
\newblock {\em arXiv e-prints}, pages arXiv--2012, 2020.

\bibitem{ruan2022survey}
Ludan Ruan and Qin Jin.
\newblock Survey: Transformer based video-language pre-training.
\newblock {\em AI Open}, 2022.

\bibitem{liu2021swin}
Ze~Liu, Yutong Lin, Yue Cao, Han Hu, Yixuan Wei, Zheng Zhang, Stephen Lin, and
  Baining Guo.
\newblock Swin transformer: Hierarchical vision transformer using shifted
  windows.
\newblock In {\em Proceedings of the IEEE/CVF International Conference on
  Computer Vision}, pages 10012--10022, 2021.

\bibitem{sandouka2021transformers}
Soha~B Sandouka, Yakoub Bazi, and Naif Alajlan.
\newblock Transformers and generative adversarial networks for liveness
  detection in multitarget fingerprint sensors.
\newblock {\em Sensors}, 21(3):699, 2021.

\bibitem{zhong2021face}
Yaoyao Zhong and Weihong Deng.
\newblock Face transformer for recognition.
\newblock {\em arXiv preprint arXiv:2103.14803}, 2021.

\bibitem{huang2018improved}
Cheng-Zhi~Anna Huang, Ashish Vaswani, Jakob Uszkoreit, Noam Shazeer, Curtis
  Hawthorne, Andrew~M Dai, Matthew~D Hoffman, and Douglas Eck.
\newblock An improved relative self-attention mechanism for transformer with
  application to music generation.
\newblock 2018.

\bibitem{huang2018music}
Cheng-Zhi~Anna Huang, Ashish Vaswani, Jakob Uszkoreit, Noam Shazeer, Ian Simon,
  Curtis Hawthorne, Andrew~M Dai, Matthew~D Hoffman, Monica Dinculescu, and
  Douglas Eck.
\newblock Music transformer.
\newblock {\em arXiv preprint arXiv:1809.04281}, 2018.

\bibitem{shi2019two}
Lei Shi, Yifan Zhang, Jian Cheng, and Hanqing Lu.
\newblock Two-stream adaptive graph convolutional networks for skeleton-based
  action recognition.
\newblock In {\em Proceedings of the IEEE/CVF conference on computer vision and
  pattern recognition}, pages 12026--12035, 2019.

\bibitem{misra2019mish}
Diganta Misra.
\newblock Mish: A self regularized non-monotonic activation function.
\newblock {\em arXiv preprint arXiv:1908.08681}, 2019.

\end{thebibliography}



\end{document}